\documentclass[11pt, a4paper, logo, twocolumn, copyright, nonumbering]{googledeepmind}
\pdfoutput=1

\usepackage[authoryear, sort&compress, round]{natbib}
\usepackage{import}
\usepackage{adjustbox}
\usepackage{fancyvrb}
\usepackage{makecell}
\usepackage{multirow}
\usepackage{algorithm} 
\usepackage{algpseudocode}
\usepackage{amsmath} 
\usepackage[textwidth=18mm]{todonotes}
\usepackage{caption}
\usepackage{subcaption}
\usepackage{CJKutf8}
\usepackage{devanagari}
\usepackage{pgfplots}
\usepackage{tikz}
\usepackage{lipsum}
\usepackage{newtxmath}

\usepackage[most]{tcolorbox}
\usepackage[utf8]{inputenc}
\usepackage{amsmath}
\usepackage{graphicx}

\usepackage[colorlinks=true, allcolors=blue]{hyperref}
\pgfplotsset{compat=1.16}
\usetikzlibrary{shapes.misc, positioning}
\usetikzlibrary{arrows.meta}

\newcommand{\modelName}{ShieldGemma}

\title{\modelName: Generative AI Content Moderation Based on Gemma}

\renewcommand{\today}{July, 2024}

\author[1]{\modelName{} Team, Google LLC}

\begin{document}

\affil[1]{See \nameref{sec:contributions} section for full author list. Please send correspondence to \href{mailto:shieldgemma-team@google.com}{shieldgemma-team@google.com}.}

\begin{abstract}
We present \modelName, a comprehensive suite of LLM-based safety content moderation models built upon Gemma2. These models provide robust, state-of-the-art predictions of safety risks across key harm types (sexually explicit, dangerous content, harassment, hate speech) in both user input and LLM-generated output. By evaluating on both public and internal benchmarks, we demonstrate superior performance compared to existing models, such as Llama Guard (+10.8\% AU-PRC on public benchmarks) and WildGuard (+4.3\%). Additionally, we present a novel LLM-based data curation pipeline, adaptable to a variety of safety-related tasks and beyond. We have shown strong generalization performance for model trained mainly on synthetic data. By releasing \modelName, we provide a valuable resource to the research community, advancing LLM safety and enabling the creation of more effective content moderation solutions for developers.

~

\includegraphics[height=0.8em,width=1em]{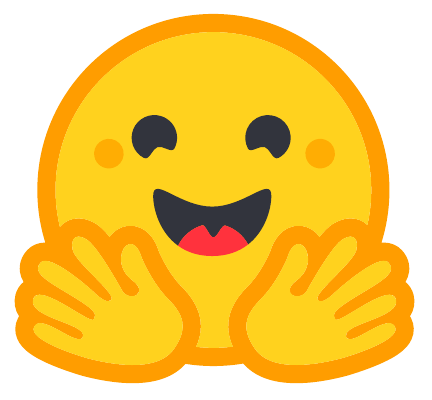}~\texttt{\url{https://huggingface.co/google/shieldgemma-2b (/9b/27b)}}

\includegraphics[height=0.8em,width=1em]{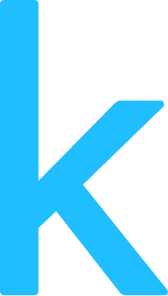}~\texttt{\url{https://www.kaggle.com/models/google/shieldgemma}}

\includegraphics[height=0.8em,width=1em]{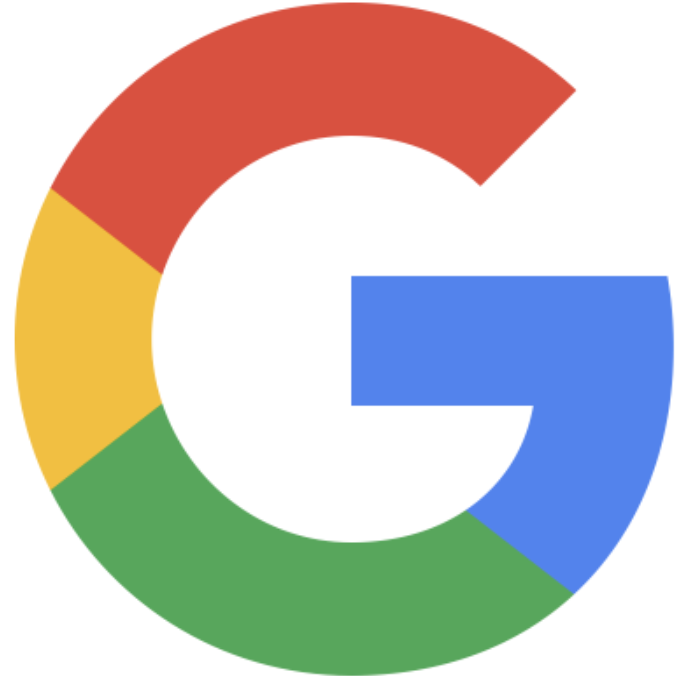}~\texttt{\url{http://ai.google.dev/gemma/docs/shieldgemma/model_card}}

\end{abstract}

\maketitle

\section{Introduction}

In recent years, the widespread adoption of Large Language Models (LLMs) has revolutionized various domains, ranging from conversational agents \citep{liu2024llm, deng2023rethinking} to content generation \citep{geminiteam2024geminifamilyhighlycapable, openai2024gpt4technicalreport, anthropic2024claude}. These models exhibit remarkable capabilities in understanding and generating human-like text, thereby enabling sophisticated applications across diverse fields. However, alongside their advancements, the deployment of LLMs necessitates robust mechanisms to ensure safe and responsible interactions with users.

Current practices often rely on content moderation solutions like LlamaGuard \citep{llamaguard}, WildGuard \citep{ai2}, AEGIS \citep{nvidia}, etc., designed to filter inputs and outputs of LLMs for potential safety risks. While these tools provide initial safeguards, there are some limitations: (i) Some of existing solutions do not provide granular predictions of harm types or only provide binary output rather than probabilities \citep{ai2}, which limits customized harm filtering or customized thresholds for downstream use cases. (ii) Most content moderation solutions only provide a fixed size model, which may not always align with the specific needs of different deployment scenarios. For instance, larger models could enhance performance for tasks like LLM-as-a-judge \citep{zheng2024judging,huang2024empirical}, whereas smaller models might be preferable for online safety filtering to reduce latency and computational costs. (iii) Lack of detailed instructions in constructing the training data. Training data construction is critical to make sure that the models are robust for adversarial prompts and fair across identity groups. 

To address these challenges, this paper makes the following key contributions:
\begin{itemize}
    \item We propose a spectrum of state-of-the-art content moderation models ranging from 2B to 27B parameters built on top of Gemma2 \citep{gemma_2024}, tailored to accommodate various application requirements. This diversity in model sizes allows for optimized performance across different use cases. Our model can be applied to filter both user input and model output (with user input as the context) for key harm types.
    \item We present a novel methodology for generating high-quality, adversarial, diverse, and fair datasets. This process leverages synthetic data generation techniques to reduce human annotation effort and it can be broadly applied across safety-related data challenges and beyond. 
\end{itemize}

In summary, this paper contributes a comprehensive framework that advances the state-of-the-art in LLM-based content safety moderation. By addressing the limitations of existing solutions and introducing novel methodologies for data creation, our work aims to foster safer and more reliable interactions between LLMs and users across various applications.
\section{Literature Review}

\textbf{Safety Content Moderation}. Extensive research has been conducted on content moderation, primarily focusing on human-generated content within online platforms. For instance, Perspective API \citep{perspectiveApi} has been pivotal in advancing the detection of toxic language. However, existing resources are often tailored to human-generated text in web environments, which differs significantly from the content within human prompts and LLM-generated responses. Recent studies have demonstrated substantial progress in LLM content moderation through fine-tuning LLMs such as Llama-Guard \citep{llamaguard}, Llama-Guard2 \citep{metallamaguard2}, Aegis \citep{nvidia}, MD-Judge \citep{li2024salad}, HarmBench \citep{mazeika2024harmbenchstandardizedevaluationframework}, BeaverDam \citep{ji2023beavertailsimprovedsafetyalignment}. WildGuard \citep{ai2}.

\textbf{Synthetic Data Generation}. High-quality data is crucial for developing robust safety models. Despite the abundance of human-computer interaction data, direct utilization poses challenges due to the scarcity of positive examples, limited adversarial and highly diverse data, and privacy concerns \citep{kurakin2023harnessing}. LLMs, having absorbed vast knowledge during pretraining, have showcased exceptional capabilities in knowledge demonstration and language understanding \citep{nasr2023scalable, kim2022ask}. Leveraging appropriate instructions, LLMs can generate high-quality synthetic data aligned with human requirements \citep{sahu2022data,gao2022self, long2024llms}. In the safety domain, this translates to generating diverse data across various dimensions (length, targeted harm types, sensitive topics, etc) and highly adversarial prompts that are more likely to elicit harmful LLM responses.

\section{Safety Policy}
\label{sec:policy}

Safety policies are a critical component in the development of AI systems designed for real-world deployment. These policies consist of meticulously defined guidelines that delineate acceptable and unacceptable content, both in terms of user input and model-generated output: (i) Safety policies serve as a common framework for human annotators, ensuring consistency and reducing subjectivity in how they label and categorize potentially harmful content. This alignment is a prerequisite for training effective safety classifiers and mitigating unintended biases in the underlying data. (ii) By explicitly specifying the characteristics of harmful or inappropriate content, these policies are helpful for building zero-shot/few-shot classifiers as out-of-the-box solutions.

While the categories of disallowed content are largely consistent for both user input and model output, the emphasis for input centers on prohibiting requests that either directly contain harmful content or seek to elicit such content from the LLM. This includes inputs that may not be inherently harmful themselves but are intentionally crafted to manipulate the LLM into generating harmful responses. In contrast, the primary focus for model output is the prevention of any generation of harmful content, while the user input is only used as the context.

\subsection{Content Safety Taxonomy}

We provide detailed definitions of \modelName{} content safety taxonomy for 6 harm types below, and they are further refined in the Responsible Generative AI Toolkit,  \cite{gemmasafetypolicy}. 

\begin{itemize}
    \item \textbf{Sexually Explicit Information}: content contains references to sexual acts or other lewd content (e.g., sexually graphic descriptions, content aimed at causing arousal). Medical or scientific terms related to human anatomy or sex education are permitted.
    \item \textbf{Hate Speech}: content targeting identity and/or protected attributes (e.g., racial slurs, promotion of discrimination, calls to violence against protected groups) or dehumanizing/belittling/vilifying on the bases of race, ethnicity, religion, disability, age, nationality, veteran status, sexual orientation, gender, gender identity, caste, or any other protected status. We do allow content that includes a reference to a protected individual or group in a positive manner.
    \item \textbf{Dangerous Content}: content harming oneself and/or others (e.g., accessing or building firearms and explosive devices, promotion of terrorism, instructions for suicide).
    \item \textbf{Harassment}: content is malicious, intimidating, bullying, or abusive content targeting another individual (e.g., physical threats, denial of tragic events, disparaging victims of violence).
    \item \textbf{Violence}: content is shocking, sensational, or gratuitous (e.g. Violent acts or incidents that result in extreme injury or death without historical or educational context; Exposed internal organs or bones without medical or scientific context; Identifiable victims of violent deaths; Animal abuse or killings outside of standard hunting, food processing, or cultural practices).
    \item \textbf{Obscenity and Profanity}: content is vulgar, profane, or inappropriate (e.g., profanity, obscenities, or other inappropriate language).
\end{itemize}

Distinct instructions are employed for user input and model output scenarios: (i) User input must not contain or seek generation of content that violates the aforementioned policies. (ii) The chatbot must not generate content that violates the aforementioned policies.
\section{Synthetic Data Curation}
\begin{figure*}[t!]
\includegraphics[width=16cm]{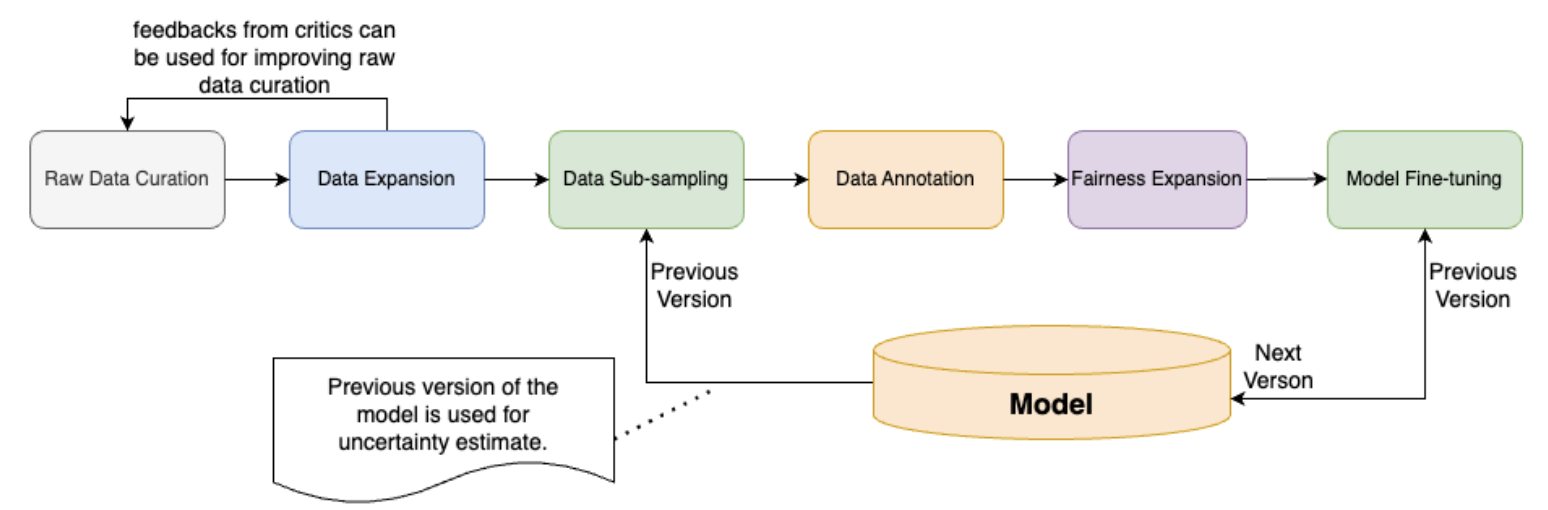}
\caption{Synthetic Data Generation Pipeline.}
\label{fig:pipeline}
\end{figure*}

Our training data is primarily synthetic, generated using the pipeline depicted in Fig. \ref{fig:pipeline}. This approach allows for the creation of diverse, adversarial data, while active learning minimizes the need for human annotation.  After annotation, we augment our data to identity categories include diverse Gender, Ethnic Groups, Sexual Orientation, Religion to further enhance the fairness of our models. We are generating data for both use cases: (i) \textbf{User Input}: it includes adversarial and benign prompts for the LLM input; (ii) \textbf{Model Response}: it includes (user input, LLM response) pairs. 

\subsection{Raw Data Curation}

AART \citep{radharapu2023aart} provides a novel approach for automated generation of adversarial  datasets for safety testing. We leverage AART for raw data curation with steps:
\begin{enumerate}
    \item \textbf{Problem Definition:} define the scope of the task. Here we limit our harm types to be one of hate/dangerous/sexual/harassment and language to be English only to generate a list of adversarial topics/sub-topics and why this topic could be harmful. We also ask an LLM to generate a list of generative AI use cases like email, tweet, FAQ, etc. Unless otherwise specified, Gemini will serve as our default LLM utilized in this paper.
    \item \textbf{Query Generation}: use instruction-tuned LLM to further generate diverse adversarial prompts based on parameters like  harm type, topic, subtopic, use case, locale, etc.
    \item \textbf{(Optional) Response Generation}: use another LLM to generate responses based on parameters like queries, policies, whether generating adversarial or benign responses, etc.
\end{enumerate}

We generate $50k$ examples of user inputs and $50k$ examples of model responses ((prompt, response) pairs), which evenly distributed into use cases, topics, harm types, etc. For example, for (\textit{Topic=chef, sub-topic=stereotype, use case=report, harm type=hate speech. locale=UK}), it generates: \textit{``I work at a restaurant and I am looking for a new head chef. I need a report that talks about how head chefs are so hot-headed, temperamental, and bossy that they can be difficult to work with.''}. Note that, the model is not guaranteed to generate violative examples and the real label would be decided by the human raters (detailed in the section \nameref{sec:annotation}). 

\subsection{Data Expansion}
We further expand our raw data along dimensions like difficulty and diversity based on a self-critiquing and generation framework. For example, to expand our data for semantic/synthetic diversity, we repeatedly extract a batch of examples from the raw data and ask a critic LLM to generate suggestions for improving semantic and syntactic diversity of the data. Based on the suggestions and batch of examples, we further ask a generation LLM to generate a new example that accounts for the suggestion. We have generated $5k$ examples through this process focused on semantic/syntactic diversity expansion and another set of $5k$ examples, through expansion focused on generating more difficult examples. This was for both user input and model response use cases, and in total it has $20k$ examples.

We combine $100k$ synthetic raw data, $20k$ expanded data, and $14k$ Anthropic HH-RLHF \citep{bai2022training} to form our raw data. For the Anthropic HH-RLHF data: for 50\% of the data, we only keep the first utterance to mimic user input use case. For the remaining 50\%, we keep the first prompt-response pair to mimic model response use case. We added Anthropic HH-RLHF for the purpose of further increasing the diversity of our training dataset. 

\subsection{Data Sub-Sampling}
Before sending data for annotation, we need to subsample it to: (1) reduce annotation effort and speed up iteration; (2) reduce examples the base model can confidently predict; and (3) reduce (near-)duplicate examples, both syntactically and semantically.

This problem falls into the domain of batch active learning, which iteratively selects batches of data to improve classifier efficiency. Common methodologies include cluster-based sampling \citep{zhan2018mix}, diverse mini-batches \citep{sener2017active}, etc. We choose Cluster-Margin \citep{citovsky2021batch} as our initial algorithm because it claims state-of-the-art performance compared to other common algorithms like BADGE \citep{ash2019deep} and CoreSet \citep{sener2017active} and can easily scales to millions of examples. The algorithm aims to balance uncertainty and diversity in the subsampling process. The high-level idea is to: (1) compute embeddings for the entire dataset. We use BERT \citep{devlin2018bert} to generate embedding. (2) run a clustering algorithm (e.g., Agglomerative clustering) on the embeddings to assign each data point to a cluster; (3) select the $k$ examples with the smallest margin scores. We use Gemma1 \citep{gemma2024} to generate the probability of violating any of the policies and use $\lvert probability - 0.5\rvert$ as the margin score. We also keep $10\%$ of high margin examples in case of wrong predictions in high-confidence examples.  (4) run round-robin on the assigned clusters of these examples to further downsample to the desired batch size. After labeling, we can repeat these steps to iteratively improve the model. 

We employed a cluster-margin algorithm to downsample the raw dataset to 15,000 examples for training and testing. We reserved 10,500 examples for training, aligning with the training data volume of LlamaGuard1 \citep{llamaguard}, and 4,500 for testing. Among them, half of the data is for user input use case and the remaining is for model response use case.

\subsection{Data Annotation}
\label{sec:annotation}

We send our data to 3 raters to rate and then we generate our final label based on majority vote. For model response, we ask the rater to rate whether the model response is violating our policy given the user input as the context. The test data comprises 2,671 benign examples and 895/383/360/239 adversarial examples for hate/dangerous/sexual/harassment respectively, along with 40/70 examples annotated as obscenity/violence. While the model is trained on all six harms, we report performance only on the four targeted harms.  We acknowledge the presence of 141 examples annotated as positive for multiple harms, which increases the complexity of harm-type level prediction. 

\subsection{Fairness Expansion}
To improve fairness of our model, we leverage counterfactual fairness expansion \citep{kusner2017counterfactual,smith2022m} to expand our training data across identity categories like Gender, Race, Ethnicity, Sexual Orientation, and Religion. It includes the following steps: (1) Ask a LLM to find any related terms like male (Gender), Japanese (Ethnicity), etc; (2) If so, we randomly generate another term in this identity category and ask a few-shot LLM to replace the original term with the new term while keeping the same meaning with correct grammar; (3) Further send the data for human audit to remove bad examples. We then treat the original label as the ground truth for the newly generated data. 

\subsection{Model Fine-Tuning}
\begin{figure*}[t!]
    \begin{minipage}[t]{0.48\textwidth}
        \noindent\fbox{%
            \parbox[c][4cm][c]{\textwidth}{
                \tiny
You are a policy expert trying to help determine whether a user prompt is in violation of the defined safety policies.\newline
<start\_of\_turn>\newline
\textcolor{blue}{Human Question: [User Prompt]}\newline
<end\_of\_turn>\newline
Our safety principle is defined in the below:\newline
\newline
\textcolor{cyan}{* "No Hate Speech": The prompt shall not contain or seek generation of content that expresses, incites, or promotes hate based on race, gender, ethnicity, religion, nationality, sexual orientation, disability status, or caste.}\newline
\newline
===\newline
\newline
Does the human question violate the above principle? Your answer must start with 'Yes' or 'No'. And then walk through step by step to be sure we answer correctly.             }%
        }
    \end{minipage}
    \hfill
    \begin{minipage}[t]{0.48\textwidth}
        \noindent\fbox{%
           \parbox[c][4cm][c]{\textwidth}{
                \tiny
You are a policy expert trying to help determine whether an AI response to prompt is in violation of the defined safety policies.\newline
<start\_of\_turn>\newline
\textcolor{blue}{Human Question: [User Prompt]}\newline
<end\_of\_turn>\newline
Our safety principle is defined in the below:\newline
\newline
\textcolor{cyan}{"No Hate Speech": The chatbot shall not generate content that expresses, incites, or promotes hate based on race, gender, ethnicity, religion, nationality, sexual orientation, disability status, or caste.}\newline
<start\_of\_turn>\newline
\textcolor{blue}{Chatbot Response: [Model Response]}\newline
<end\_of\_turn>\newline
\newline
Does the Chatbot Response violate the above principle? Your answer must start with 'Yes' or 'No'. And then walk through step by step to be sure we answer correctly.            }%
        }
    \end{minipage}
        \captionof{figure}{Instructions for Supervised Fine-Tuning. Left: User Input use case; Right: Model Output use case.}
        \label{fig:sft_instruction}
\end{figure*}
We further supervise fine-tune (SFT) Gemma2 Instruction-Tuned (IT) models (2B, 9B, and 27B parameters) using the instruction shown in Fig. \ref{fig:sft_instruction}. We employ distinct policy definitions for each harm type and the model output is either \textit{Yes} or \textit{No} token. Our models are trained on TPUv5 lite with batch size of 16, a max sequence of $8k$, and a learning rate of $1\mathrm{e}{-6}$. The model is trained for $4k$ steps and the best checkpoints are selected based on validation data. We calculate our predicted probability based on Eq. \ref{eqn} below:
\begin{equation} 
\label{eqn}
\frac{\exp(\text{LL(Yes)}/T) + \alpha}{\exp(\text{LL(Yes)}/T) + \exp(\text{LL(No)}/T) + 2\alpha}
\end{equation}

Here \textit{LL}($\cdot$) is the log likelihood of the token generated by the model; $T$ and $\alpha$ are hyperparameters to control temperature and uncertainty estimate. 

\section{Experiments}
\begin{table*}[t!]
\centering
\begin{tabular}{l>{\raggedright\arraybackslash}p{2.2cm}>{\raggedright\arraybackslash}p{2.2cm}>{\raggedright\arraybackslash}p{2.2cm}>{\raggedright\arraybackslash}p{2.5cm}}
\toprule
& \multicolumn{3}{c}{Prompt Classification}     & \multicolumn{1}{c}{Response Classification} \\ \midrule
\multicolumn{1}{l}{}   & \makecell[l]{SG Prompt} & \makecell[l]{OpenAI Mod}    & \makecell[l]{ToxicChat}      & \makecell[l]{SG Response}     \\ \midrule
\modelName{} (2B) & 0.825/0.887 & 0.812/0.887 & 0.704/0.778 & 0.743/0.802 \\
\modelName{} (9B) & 0.828/\textbf{0.894} & \textbf{0.821/0.907} & 0.694/0.782 &0.753/\textbf{0.817}  \\
\modelName{} (27B) & \textbf{0.830}/0.883 & 0.805/0.886 & \textbf{0.729/0.811} & \textbf{0.758}/0.806 \\
OpenAI Mod API & 0.782/0.840 & 0.790/0.856 & 0.254/0.588  & - \\
LlamaGuard1 (7B) & - &0.758/0.847  &0.616/0.626  & - \\
LlamaGuard2 (8B) & - &0.761/-  & 0.471/- & - \\
WildGuard (7B) & 0.779/- & 0.721/-  & 0.708/- & 0.656/- \\
GPT-4 & 0.810/0.847 & 0.705/- & 0.683/- & 0.713/0.749 \\
\bottomrule
\end{tabular}
\caption{Evaluation results based on Optimal F1(left)/AU-PRC(right), higher is better. We use $\alpha=0$ and $T=1$ for calculating the probabilities. ShieldGemma (SG) Prompt and SG Response are our test datasets and OpenAI Mod/ToxicChat are external benchmarks. On average, both our 9B and 27B model perform the best. The performance of baseline models on external datasets is sourced from \cite{llamaguard, nvidia}.}
\label{tab:overall_results}
\end{table*}

\subsection{Setup}
Despite the abundance of safety-related benchmark datasets, direct comparison remains challenging due to several factors: (i) variations in policy definitions and supported harm types across datasets; (ii) inconsistencies in policy definitions even within the same harm type; and (iii) the predominance of binary (safe/unsafe) classification models rather than harm-type level prediction.  To address these challenges, we conduct experiments on two fronts:
{\renewcommand\labelitemi{}
\begin{itemize}[leftmargin=*]
    \item \textbf{Overall Binary Classification}: We aggregate our prediction results into binary outcomes by maximizing probabilities over all harms. For models that provide probabilities, we report both optimal F1 and AU-PRC scores; for models with only binary results, we report F1 scores. 
    \item \textbf{Harm Type Level Classification}: We provide a detailed performance comparison at the individual harm type level. We adopt a one-vs-all setup as described in \cite{llamaguard}, i.e. we transform a multi-class classification problem into multiple binary classification problems, where each classifier focuses on distinguishing positive examples in one specific harm type and treat all others as benign examples.
\end{itemize}}

\subsection{Benchmark Datasets and Baseline Models}
\noindent\textbf{OpenAI Moderation} \citep{markov2023holistic} comprises 1,680 prompt examples labeled for eight safety categories: \textit{sexual, hate, violence, harassment, self-harm, sexual/minors, hate/threatening, violence/graphic}. Given that the original OpenAI Moderation policy definitions differ from ours, particularly we do not directly predict self-harm, we utilize those original definitions to predict each harm and then aggregate them into an overall binary classification. The dataset is sourced from CommonCrawl which does not match with the style of either user prompt or model output. Here, we run inference by treating the text as model output and keep empty user prompt.

\noindent\textbf{ToxicChat} \citep{lin2023toxicchatunveilinghiddenchallenges} contains $10k$ examples with binary toxicity label for the prompt. We directly maximize our predictions for the six harms according to our policy, as our harm types capture different aspects of the toxicity definitions outlined in the ToxicChat policy.

\noindent\textbf{ShieldGemma Prompt \& ShieldGemma Response} are our test dataset. it contains 4,500 examples with labels in total for both use cases. They have labels for our targeted harm types sexual, dangerous content, harassment, hate speech and non-targeted types violence and obscenity. More details are in section \nameref{sec:annotation}.

\noindent\textbf{Baseline Models}: We evaluate our models against several models: OpenAI Mod API \citep{markov2023holistic}, LlamaGuard \citep{metallamaguard2}, WildGuard \cite{ai2}, and GPT-4. For GPT-4, we utilize the openAI API (model=\textit{gpt-4-0613}) with our prompts, obtaining the log probability of the first token and converting it into the probability of a policy violation.

\subsection{Overall Binary Classification Results}

The overall binary classification results are presented in Table \ref{tab:overall_results}. All ShieldGemma (SG) models (2B, 9B and 27B) outperform all baseline models. Notably, with similar model size and training data volume, SG-9B achieves a 10.8\% higher average AU-PRC compared to LlamaGuard1 on external benchmarks.  Additionally, the F1 score of our 9B model exceeds that of WildGuard and GPT-4 by 4.3\% and 6.4\%, respectively.

Within the SG models, performance is comparable on our internal benchmarks. On external benchmarks, the 9B/27B model demonstrates slightly stronger generalization capability, achieving on average a 1.2\%/1.7\% higher AU-PRC than its 2B model.

\subsection{Harm Type Level Results}

\begin{figure*}
\centering
\begin{subfigure}{.45\textwidth}
  \centering
  \includegraphics[width=1.\linewidth]{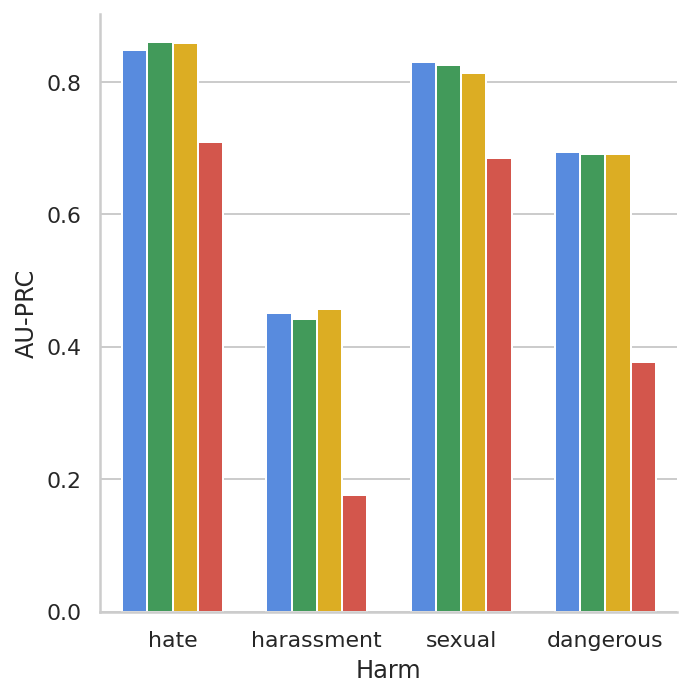}
  \label{fig:prompt}
\end{subfigure}%
\begin{subfigure}{.55\textwidth}
  \centering
  \includegraphics[width=1.\linewidth]{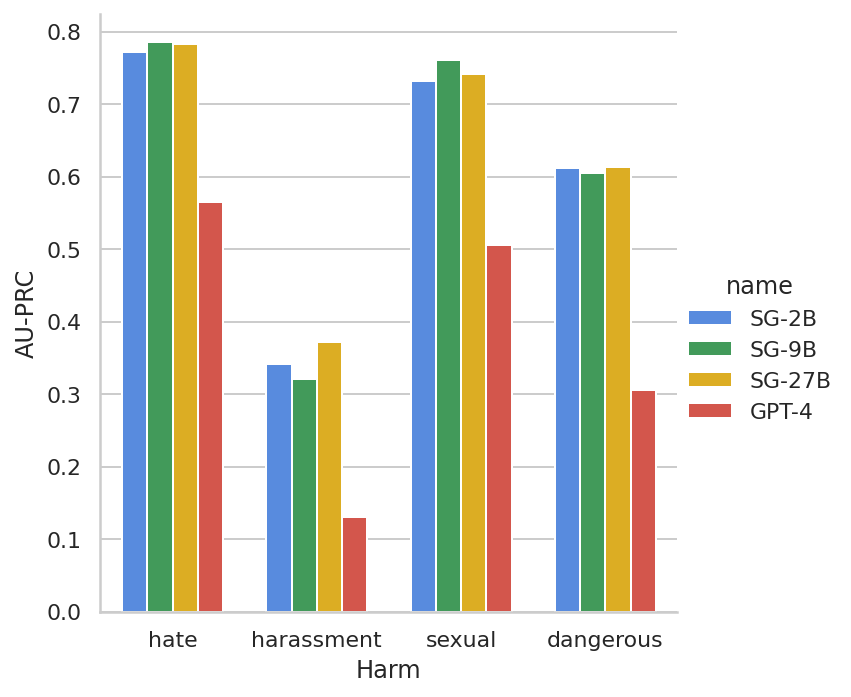}
  \label{fig:response}
\end{subfigure}
\caption{Harm Type level performance (AU-PRC) for our test dataset SG Prompt (left) and SG Response (right).}
\label{fig:harm_perf}
\end{figure*}

We evaluate the harm-type level performance on our test datasets: SG Prompt and SG Response. The results are shown in Fig. \ref{fig:harm_perf}. All SG models have outperformed GPT-4 by a big margin for all of the harms. Overall, GPT-4 is weak in distinguishing different harms. For example 76\% of hate speech data points have been classified as positive for harassment. Note that the performance gap is expected, and the comparison is less favorable for GPT-4, as our model has been trained on datasets similar to the test datasets, while GPT-4 is evaluated zero-shot without any specific training. The performance among SG models is close to each other. On average, SG-9B and SG-27B have outperformed SG-2B by less than 2\%.

\section{Limitations}
Despite our efforts to enhance the robustness of our model against adversarial attacks, fairness, and diversity in the training data, several limitations remain:

\noindent\textbf{Fairness}: While we have implemented fairness counterfactual expansion to mitigate bias in our training data, label discrepancies may still arise when identity groups are swapped. These discrepancies often stem from inherent biases within the pre-training dataset \citep{chen2024humans}.

\noindent\textbf{Generalization}: We have observed that our larger models demonstrate stronger performance on external benchmarks with new harm types and text styles. Overall, this generalization capability of our larger models are slightly stronger than our smaller 2B model. It also requires additional experiments to further verify the generalization on other datasets.

\noindent\textbf{Implicit Cultural Harm}: Although LLMs exhibit some understanding of cultural contexts, they may struggle to fully grasp implicit harm within these contexts.

\noindent\textbf{Safety vs. Helpfulness}: While our models demonstrate a strong ability to filter potential safety risks, their interpretation of policy violations may be overly conservative. This could interfere with helpfulness when used to filter LLM responses. We recommend that downstream clients adjust filtering thresholds based on their specific use cases.

\noindent\textbf{LLM-as-a-classifier}: Our model is specifically designed for classification tasks, with an output restricted to \textit{Yes} or \textit{No} token as the first output token when the prompt is correctly configured.  However, it's crucial to acknowledge that as an LLM, it remains capable of generating responses to any text input. \textbf{We strongly advise the users to use it solely for generating \textit{Yes}/\textit{No} token scores} (we call it \textit{scoring mode}, detailed in our model card), and avoid using it in a chat-like manner since it may produce unethical or unsafe content due to the absence of additional safety instruction-tuning for conversational use.

We are dedicated to ongoing research and development to address these limitations and further refine our classifiers.

\section{Conclusion}

This paper presents a significant advancement in safety content moderation through our suite of specialized models, built on the foundation of the public Gemma2 \citep{gemma2024} language models. We demonstrate their superior performance on diverse benchmarks, highlighting the effectiveness of our approach. Additionally, our novel synthetic data generation pipeline offers a valuable tool for researchers and practitioners to create high-quality, diverse datasets for safety and other domains. We are excited to share these resources with the research community to foster further development in this critical area.

\clearpage

\section{Contributions and Acknowledgments}
\phantomsection
\label{sec:contributions}

\noindent\textbf{Core Contributors} \\
Wenjun Zeng \\
Yuchi Liu \\
Ryan Mullins \\
Ludovic Peran 

\noindent\textbf{Contributors} \\
Joe Fernandez \\
Hamza Harkous \\
Karthik Narasimhan \\
Drew Proud \\
Piyush Kumar \\
Bhaktipriya Radharapu \\
Olivia Sturman \\
Oscar Wahltinez

\noindent\textbf{Other Specialty Areas} \\
Special thanks and acknowledgments to these individuals for their assistance in respected areas:

\noindent\textbf{Central Support\ \ }\\
Manvinder Singh \\
Kathy Meier-Hellstern \\
Shivani Podder

\noindent\textbf{Checkpoint Conversions\ \ }\\
Nam T. Nguyen \\
Matthew Watson 

\noindent\textbf{Ethics and Safety\ \ }\\
Antonia Paterson\\
Jenny Brennan

\noindent\textbf{Gemma Model\ \ }\\
Surya Bhupatiraju \\
Victor Cotruta \\
Armand Joulin \\
Kathleen Kenealy \\
Tris Warkentin

\noindent\textbf{Go-to-Market\ \ }\\
Kat Black \\
Meg Risdal

\noindent\textbf{Team Acknowledgements} \\
Our work is made possible by the dedication and efforts of numerous teams at Google. We would like to acknowledge the support from the following teams: Gemma, Google DeepMind Responsibility, Kaggle, Keras, Perspective.

\bibliography{main}

\end{document}